\documentclass{article}

\usepackage{graphicx}
\usepackage{subcaption}
\usepackage{amsmath}
\usepackage{amsfonts}
\usepackage{amssymb}
\usepackage{mathtools}
\usepackage{enumerate}
\newtheorem{definition}{Definition}[section]
\newtheorem{remark}{Remark}[section]

\newtheorem{problem}{Problem}[section]

\usepackage{booktabs}
\usepackage{tabularx}
\usepackage{multirow}
\usepackage{makecell}
\usepackage{enumitem} 
\usepackage{bm}
\usepackage{adjustbox}

\usepackage[final]{corl_2026} 
\title{BarrierFormer: Transformer-Guided Predictive Barrier Enforcement for Safe Robot Control}

\author{
  Anandsingh Chauhan\\
  Arizona State University \\
  \texttt{achauh67@asu.edu}
  \And
  Kunal Garg \\
  Arizona State University \\
  \texttt{kgarg24@asu.edu}
}

\begin{document}
\maketitle

\begin{abstract}
Control barrier functions (CBFs) have become one of the most popular tools for encoding and enforcing state constraints in safety-critical robotics. Standard CBF approaches are inherently myopic in nature as they enforce safety only at the current time step. Consequently, the system can be driven toward the boundary of the safe set where no feasible safe control exists at a future timestep. Model predictive control (MPC) based approaches address this by enforcing state constraints over a receding horizon. However, such approaches generally require the model to be known for solving a constrained optimization problem at every step, which is computationally expensive for real-time deployment. We propose BarrierFormer, a barrier-supervised transformer framework that addresses these limitations by encoding rollout-level CBF constraints in learning a model-free safe policy. A causal transformer encodes observation-action history, autoregressively generates a predictive rollout through the dynamics head to replace the model, and provides a residual correction to a nominal controller through the action head to replace the online computation. A barrier critic operating on local observations evaluates CBF constraint violations along this rollout, and a safety teacher computes barrier-consistent actions satisfying these constraints as direct supervision targets for the learned control policy. During inference, the policy maps observation-action history to control actions without any online optimization or model knowledge, enabling real-time model-free predictive safety enforcement. Evaluations across linear and nonlinear, 2D and 3D dynamical systems for safe goal-directed navigation demonstrate that BarrierFormer outperforms existing reinforcement learning (RL)-based, diffusion-based, MPC-based, and transformer-based approaches in safety rate and inference latency.\footnote{Project website: \href{https://safe-robotics-group-at-asu.github.io/BarrierFormer-website/}{https://safe-robotics-group-at-asu.github.io/BarrierFormer-website/}}
\end{abstract}



\section{Introduction}
Safe robot navigation requires an agent to reach a goal while avoiding obstacles~\cite{So-RSS-23, palanisamy2020multi,marl-safe}. This reach-avoid objective couples two aspects: the controller must make progress toward the goal while avoiding actions that lead to future safety violations~\cite{collisionavoid}. Control barrier functions (CBFs) provide a principled framework for the safety component by converting set-invariance requirements into constraints on the control input~\cite{ames2017control, ames2019control, dimitrasafe}, minimally modifying a nominal controller to be safe. An online control input is computed using a single-step quadratic program (QP) to enforce these constraints, but this pointwise formulation is myopic: the barrier condition is evaluated only at the current step and does not constrain how the closed-loop trajectory evolves~\cite{belta, myopic-cbf, differentiable-cbf}. An action may therefore be instantaneously safe while still driving the system toward states where the barrier constraint becomes infeasible and no safe recovery action remains~\cite{belta, zeng2021safety}. Model predictive control (MPC) and discrete-time CBF-MPC mitigate this by enforcing safety over predicted trajectories~\cite{zeng2021safety, mpc_cbf}, but require model knowledge for solving constrained optimization problems online, making real-time deployment computationally expensive~\cite{mpc-computation, mpcsurvey}.

The challenge is further compounded when the agent operates under partial observations. The agent observes only its own state and LiDAR readings within a fixed sensing radius, with obstacle geometry beyond that range entirely unknown at decision time~\cite{kim2025visibility, harms2024neural,lidarcbf}. A single observation is therefore insufficient to characterize the safety of the current state: the agent may be \textit{currently} safe but on a trajectory toward an obstacle that has not yet entered its sensing range. Certifying safety, therefore, requires generating a predictive rollout from the current state to anticipate future constraint violations. Diffusion-based approaches incorporate safety through predefined CBFs via iterative optimization at inference time~\cite{safediffuser, cobl-diffusion}. Transformer-based policies are better suited for this setting, leveraging self-attention to aggregate temporal information across observation-action histories~\cite{pact, conbat}; jointly predicting actions and future states through a shared transformer further improves policy performance~\cite{dreamzero}. However, incorporating safety filters or constrained optimization at deployment~\cite{conbat, barriernet, so2024train} sacrifices inference-time efficiency. The key question is therefore whether safety constraints evaluated over predicted rollouts can be encoded directly into policy parameters, such that no online optimization is needed at deployment.

In this work, we use a barrier-supervised transformer framework that directly encodes rollout-level CBF constraints into policy parameters via training-time supervision. A causal transformer encodes observation-action history and jointly outputs a residual correction to a nominal controller and predicted state increments, which are used to autoregressively generate a predictive rollout (see Fig.~\ref{fig:algo-structure-left}). A barrier critic evaluates CBF constraint violations along this rollout, and a safety teacher computes barrier-consistent corrections as direct supervision targets for the policy. Since horizon-level safety reasoning is encoded in the policy parameters via training-time barrier supervision, the policy maps observation-action histories directly to control actions, with no online optimization at inference.

The contributions of this work are as follows. {First, we propose BarrierFormer, a model-free framework for safety enforcement that encodes observation-action histories and incorporates barrier constraints over a future horizon into safe policy learning. Second, we replace the online computation of the safe input over a horizon with a learned policy, leading to a faster runtime inference compared to existing predictive methods~\cite{mpc_cbf, safediffuser}. Third, BarrierFormer uses past observation-action history, where observation includes the robot state and local LiDAR measurements, without access to a global map, to predict state increments and reconstruct future observations, thereby enforcing safety with partial environmental information.} We evaluate BarrierFormer on linear and nonlinear robot dynamics against RL-based, diffusion-based, MPC-based, and transformer-based baselines, demonstrating improved safety rate, faster inference, and zero-shot generalization to larger and more cluttered environments.

\section{Related Work}
\textbf{Optimization-based.} CBF-QP controllers enforce safety by solving a quadratic program at each timestep that minimally modifies a nominal control input to satisfy the barrier condition~\cite{ames2017control, ames2019control}. While real-time efficient, this formulation is myopic and can drive the system toward states where the barrier constraint becomes infeasible~\cite{belta}. Discrete-time CBF-MPC methods address this by enforcing barrier constraints over a receding horizon~\cite{zeng2021safety, mpc_cbf}, but the resulting optimization is generally nonconvex and requires accurate dynamics models and full state estimates, making real-time deployment expensive.

\noindent\textbf{Learning-based.} CBFs have been integrated with RL to improve constraint satisfaction through action projection or rollout filtering during training~\cite{cbfrl, cheng2019end, emam2022safe}. {These methods often require a safety filter or action projection at inference to improve safety, instead of using the policy output directly.} Other approaches learn safety certificates directly: neural CBF methods construct value-function-based safety filters~\cite{so2024train}, and BarrierNet embeds differentiable quadratic programs into neural controllers~\cite{barriernet}. Both enable end-to-end training of safety-aware policies but retain a safety filter or constrained optimization at inference.

\noindent\textbf{Generative Models for Robot Control.} Diffusion-based planners incorporate safety by embedding predefined CBFs into the denoising process via gradient projection~\cite{safediffuser, cobl-diffusion}. While these methods improve constraint satisfaction at the trajectory level, they rely on iterative denoising during inference and are not well-suited for real-time control. Transformer-based policies leverage self-attention to aggregate temporal information across observation-action histories for history-conditioned control~\cite{pact, conbat, transformermpc}. PACT does not enforce safety constraints~\cite{pact}, while TransformerMPC retains online optimization at inference~\cite{transformermpc}. ConBaT~\cite{conbat} augments a causal transformer with a learned barrier critic trained on safe and unsafe demonstrations, and at inference applies a lightweight gradient-based optimization to correct the proposed action only when {the critic predicts the next state to be unsafe}, rather than encoding safety directly into the policy parameters.

\section{Preliminaries and Problem Formulation}
\label{sec:problem_formulation}
\textbf{Notations}:~ Let $\mathbb{R}$ and $\mathbb{Z}_{\geq0}$ denote the sets of real numbers and nonnegative integers, respectively. We use $\|\cdot\|$ to denote the Euclidean norm and $[\cdot]_+$ to denote $\max(0, \cdot)$.

\textbf{System and Observation Model}:~We consider the reach-avoid problem, in which an autonomous agent safely navigates an obstacle-filled environment to reach a goal position $\mathbf{p}_{\mathrm{goal}} \in \mathbb{R}^{n_p}$~\cite{So-RSS-23}. Let $\mathbf{x}_t \in \mathcal{X} \subseteq\mathbb{R}^{n_x}$ and $\mathbf{u}_t \in \mathcal{U} \subset \mathbb{R}^{n_u}$ denote the state and control respectively at time $t \in \mathbb{Z}_{\geq 0}$, where $\mathcal{U}$ is a compact admissible control set. {Each state $\mathbf{x}_t \in \mathcal{X}$ includes a position component $\mathbf{p}_{{t}} \in \mathbb{R}^{n_p}$, where $n_p \in \{2, 3\}$ denotes the dimension of the position vector.} The agent dynamics are given by:
\begin{equation}
    \label{eq:discrete_dynamics}
\mathbf{x}_{t+1} = F(\mathbf{x}_t,\mathbf{u}_t) \coloneq f(\mathbf{x}_t)+g(\mathbf{x}_t)\mathbf{u}_t.
\end{equation}

The agent has no prior knowledge of obstacle locations and perceives its environment through $n_\text{rays}$ evenly spaced LiDAR rays within a fixed sensing radius $R>0$. Let \(o:\mathcal{X}\rightarrow\mathbb{R}^{n_o}\) denote the observation map. The observation at time \(t\) is
\begin{equation}
\label{eq:observation}
\mathbf{o}_t = o(\mathbf{x}_t) = \left( \mathbf{x}_t,\, \mathbf{p}_{\textrm{goal}}-\mathbf{p}_t,\,
\mathbf{y}_{t,1},\ldots, \mathbf{y}_{t,n_{\textrm{rays}}} \right) \in\mathbb{R}^{n_o}.
\end{equation}
We drop the time index t and the state argument $\mathbf{x}$ when clear from context. 
For each $j\in\{1,\ldots,n_{\textrm{rays}}\}$, $\mathbf{y}_{t,j}\in\mathbb{R}^{n_p+2}$ denotes the feature vector of the $j$-{\textrm{th}} LiDAR ray:
\( \mathbf{y}_{t,j} = \left( m_{t,j}, \bar d_{t,j}, \mathbf{v}_{t,j} \right), \) where $m_{t,j}$ is the hit indicator, $\bar d_{t,j}$ denotes the return distance of the $j$-th LiDAR ray, normalized by the sensing radius $R$, and $\mathbf{v}_{t,j}\in\mathbb{R}^{n_p}$ is the ray direction. 

\textbf{Safety Specification}:~Let $r \in (0, R)$ denote the agent radius and {$\mathcal{O} = \{\mathcal{O}_1, \ldots, ,\mathcal{O}_M\}$} the set of obstacles, with {$\mathcal{O}_i\subset\mathbb{R}^{n_p}$ for each $i\in\{1,\ldots,M\}$.} The prescribed safe set $\mathcal S\subset\mathcal X$ and avoid set $\mathcal A\subset\mathcal X$ are defined as:
\begin{equation}
\label{eq:safe_set}
\mathcal{S} = \left\{ \mathbf{x}\in\mathcal{X} \mid d(\mathbf{p},\mathcal{O}_i)\geq r,\;
\forall i\in\{1,\ldots,M\} \right\}, \qquad
\mathcal{A}=\mathcal{X}\setminus\mathcal{S}.
\end{equation}
{where $\mathbf{p}\in\mathbb{R}^{n_p}$ is the position component of $\mathbf{x}$, and $d(\mathbf{p},\mathcal{O}_i) = \min_{\mathbf{y}\in\mathcal{O}_i} \|\mathbf{p}-\mathbf{y}\|$ is the distance from $\mathbf{p}$ to the obstacle $\mathcal{O}_i$. The \textit{safety} requirement is that the system trajectories remain in the set $\mathcal{S}$ at all times, i.e., $\mathbf{x}_t\in\mathcal{S}$ for all $t$. A set $\mathcal{C}\subseteq\mathcal{X}$ is \emph{control invariant} for the system in~\eqref{eq:discrete_dynamics} if, for every $\mathbf{x}\in\mathcal{C}$, there exists $\mathbf{u}\in\mathcal{U}$ such that $F(\mathbf{x},\mathbf{u})\in\mathcal{C}$~\cite{blanchini1999set}. A state-feedback policy $\pi:\mathcal{X}\rightarrow\mathcal{U}$ renders $\mathcal{C}$ \emph{forward invariant} if $\mathbf{x}_0\in\mathcal{C}$ implies $\mathbf{x}_t\in\mathcal{C}$ for all $t\geq0$ under $\mathbf{x}_{t+1}=F(\mathbf{x}_t,\pi(\mathbf{x}_t))$. If $\mathcal{C}\subseteq\mathcal{S}$, forward invariance of $\mathcal{C}$ implies system safety with respect to $\mathcal{S}$. A barrier function, defined below, certifies control invariance of a subset of $\mathcal{S}$~\cite{ames2017control,ames2019control,prajna2004safety}.}
%
\begin{definition}[Discrete-Time Control Barrier Function~\cite{zeng2021safety,agrawal2017}] A function {$h:{\mathbb{R}^{n_o}}\to\mathbb{R}$} is a discrete-time control barrier function (DTCBF) for the system in~\eqref{eq:discrete_dynamics} with observation model \eqref{eq:observation} on $\mathcal{C}_h:=\{\mathbf{x}\in\mathcal{X}\mid h({o(\mathbf{x})})\geq0\}\subseteq\mathcal{S}$ if there exists a continuous function $\alpha:\mathbb{R}\rightarrow\mathbb{R}$ that is strictly increasing with $\alpha(0)=0$ and $0<\alpha(s)\leq s$ for all $s>0$, such that for every $\mathbf{x}\in\mathcal{C}_h$ there exists $\mathbf{u}\in\mathcal{U}$ such that
\begin{equation}
\label{eq:dtcbf_condition}
h(\mathbf{o}^{+})-h(\mathbf{o}) \geq -\alpha\!\left(h(\mathbf{o})\right),
\end{equation}
where ${\mathbf{o}:={o}(\mathbf{x})}$ and ${\mathbf{o}^{+}:={o}(F(\mathbf{x},\mathbf{u}))}$.
\end{definition}
We use the linear function $\alpha(s)=\gamma s$ with $\gamma\in(0,1]$, yielding $h(\mathbf{o}^{+})\geq(1-\gamma)h(\mathbf{o})$
from~\eqref{eq:dtcbf_condition}. Based on the DTCBF definition, the set of barrier-admissible control inputs is defined as:
\begin{equation}
\label{eq:admissible-input}
\mathcal{K}_h(\mathbf{x}) := \left\{\mathbf{u}\in\mathcal{U} \;\middle|\;
{
h(o(F(\mathbf x, \mathbf u))-h(o(\mathbf x)) +\alpha\!\left(h(o(\mathbf x))\right) \geq 0} \right\}.
\end{equation}
Any {observation-feedback} policy $\pi:{\mathbb{R}^{n_o}}\to\mathcal{U}$ satisfying $\pi(o(\mathbf x))\in\mathcal{K}_h(\mathbf{x}),$ for all $\mathbf{x}\in\mathcal{C}_h$, renders $\mathcal{C}_h$ forward invariant~\cite{zeng2021safety,agrawal2017}.

\begin{remark}[Myopia of Pointwise DTCBF Enforcement]
\label{rem:myopia}
Condition~\eqref{eq:dtcbf_condition} constrains a single transition at a given time-step $t$, and does not guarantee that the system can remain safe in the future. Due to this myopic nature of the CBF, a control input satisfying~\eqref{eq:dtcbf_condition} at every timestep can still drive the system towards a state ${\mathbf{x}}$ where $\mathcal{K}_h({\mathbf{x}})$ is empty~\cite{belta, zeng2021safety}, making the barrier constraint infeasible. This motivates enforcing the barrier condition over a predicted rollout rather than pointwise at the current state. Similar ideas have been explored previously by combining the CBF-based approaches with MPC-like lookahead mechanisms {\cite{mpc_cbf, breeden2022predictive, katriniok-cdc, huang2025predictive}}.
\end{remark}
{The myopic nature of the CBF constraint in \eqref{eq:dtcbf_condition} can be overcome by enforcing the CBF constraint over a horizon $\{t+k\}_{k = 0}^H$ for some $H$ at each time step $t$ as:
\begin{equation}
    \label{eq:horizon_cbf_constraint}
    h\left({\mathbf{o}_{t+k+1}}\right)
    \geq
    (1-\gamma)\,
    h\left({\mathbf{o}_{t+k}}\right),
    \quad k = 0, \ldots, H-1,
\end{equation}
which requires \textit{computation} of future states per \eqref{eq:discrete_dynamics}, which in turn requires a known system model. In a model-free control design, future states must be \textit{predicted} from available information, such as historical state-action pairs. Based on this, we formulate the problem considered in this work.}


\begin{problem}
\label{prob:main}
Design a model-free, history-conditioned policy $\pi_\theta$ such that 
the closed-loop trajectories of \eqref{eq:discrete_dynamics} satisfy \textbf{safety}, i.e., the robot does not collide with any obstacle, i.e., $d(\mathbf{p}_t,\mathcal{O}_i)\geq r,\ \forall i\in\{1,\ldots,M\},\ \forall t\geq0${,} 
where safety at each {time step} is {enforced} by requiring the barrier condition to hold over the predicted rollout~\eqref{eq:horizon_cbf_constraint}; and \textbf{liveness}, i.e., $\inf_{t\geq0}\left\|\mathbf{p}_t-\mathbf{p}_{\mathrm{goal}}\right\|=0$.

\end{problem}

\section{Methodology}
\label{sec:methodology}
\textbf{BarrierFormer overview:} {In the model-free, partially observed setting, generating a predicted rollout requires using previous observations and actions to infer how the robot state evolves}. We therefore use a GPT-2-style causal transformer~\cite{radford2019language}, which conditions residual-action and state-increment predictions on the preceding finite observation--action history containing \(T\) observations and the \(T-1\) preceding actions, where \(T\) is a positive integer:
\begin{equation}
\mathcal H_t := \left( \mathbf o_{t-T+1},\mathbf u_{t-T+1}, \ldots,
\mathbf o_{t-1},\mathbf u_{t-1}, \mathbf o_t \right) \in \mathbb H_T.
\label{eq:history}
\end{equation}
At each rollout step $t$, the predicted state increment $d_{\theta_x}$ is used to obtain the next state $\hat {\mathbf x}_{t+1}$, from which the next observation $ \hat{\mathbf o}_{t+1}$ is constructed. The current rollout history $\mathcal H_t$ is advanced to $\mathcal H_{t+1}$ by discarding the oldest observation-action pair and appending the predicted action and reconstructed observation. During training, BarrierFormer uses these predictions to enforce the barrier condition over a future horizon ${t,t+1,\ldots,t+H}$, instead of only at the current time step $t$, as discussed in Remark~\ref{rem:myopia}.
\begin{figure*}[t]
    \centering
    \includegraphics[width=\linewidth]{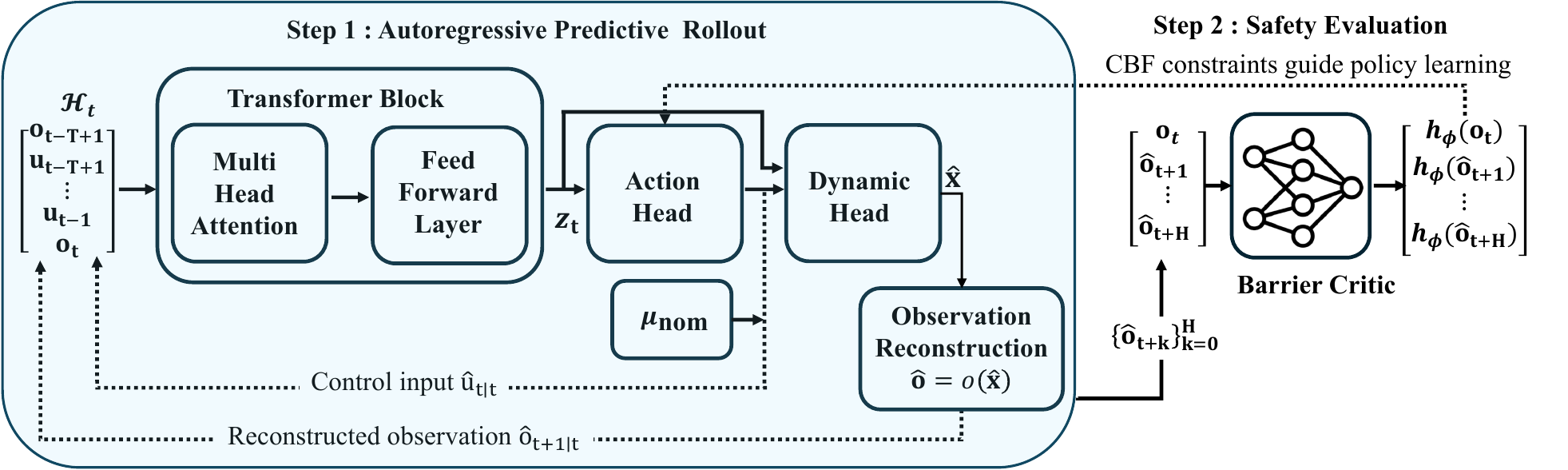}
    \caption{Overview of BarrierFormer: In Step 1, a causal transformer autoregressively generates a finite-horizon predictive rollout via the actor and dynamics heads. In Step 2, a barrier critic $h_\phi$ evaluates DTCBF constraint violations along the predicted observations.}
    \label{fig:algo-structure-left}
\end{figure*}
The proposed BarrierFormer operates in two steps, as illustrated in Fig.~\ref{fig:algo-structure-left}. In Step 1, a causal transformer encodes the observation-action history $\mathcal{H}_t$ and jointly predicts a residual correction to the action produced by a nominal control policy via the actor head and state increments via the dynamics head.  These predictions are used to autoregressively generate a finite-horizon rollout. In Step 2, a barrier critic evaluates DTCBF constraint violations along the predicted rollout, and an SQP-based safety teacher computes barrier-consistent corrections as direct supervision targets for the actor head. Step 2 operates only during training; at deployment, the transformer maps the observation-action history to control actions directly, with horizon-level safety reasoning encoded into the policy parameters through training-time barrier supervision, thereby removing the requirement of any online constrained optimization.

Each observation and action in $\mathcal H_t$ is independently mapped to a $d_h$-dimensional embedding using separate linear encoders. The resulting $2T-1$ embeddings are interleaved in temporal order to form the input token sequence. The backbone transformer $T_{\theta_T}$, parameterized by $\theta_T$, processes this sequence and produces a latent representation $\mathbf{z}_t = T_{\theta_T}(\mathcal H_t)$ that is shared between two heads. The first one is \textit{action head} $\pi_{\theta_u}$, an MLP that maps $\mathbf{z}_t$ to a residual correction $\Delta \mathbf{u}_t = \pi_{\theta_u}(\mathbf{z}_t)$, where the deployed control input is \(\mathbf{u}_t = \mu_{\mathrm{nom}}(\mathbf{x}_t) + \Delta \mathbf{u}_t\), with \(\mu_{\mathrm{nom}}\) being a nominal control policy designed to drive the robot toward the goal, {but not necessarily to satisfy the safety constraints. The second one is \textit{dynamics head} $d_{\theta_x}$, an MLP that maps $(\mathbf{z}_t, \mathbf{u}_t)$ to a predicted state increment $\Delta \hat{\mathbf{x}}_t = d_{\theta_x}(\mathbf{z}_t, \mathbf{u}_t)$, used during training to generate predictive rollouts.}
In particular, with $\hat{\mathbf{x}}_{t|t} = \mathbf{x}_t$ and $\widehat{\mathcal{H}}_{t|t} = \mathcal{H}_t$, at each step $k = 0, \ldots, H-1$, BarrierFormer computes:
\begin{align}
    \mathbf{z}_{t+k|t} &= T_{\theta_T}(\mathcal{H}_{t+k|t}), \label{eq:rollout_latent}\\
    \hat{\mathbf{u}}_{t+k|t} &= \mu_{\mathrm{nom}}(\hat{\mathbf{x}}_{t+k|t}) + 
    \pi_{\theta_u}(\mathbf{z}_{t+k|t}), \label{eq:rollout_control}\\
    \hat{\mathbf{x}}_{t+k+1|t} &= \hat{\mathbf{x}}_{t+k|t} + d_{\theta_x}( \mathbf{z}_{t+k|t}, 
      \hat{\mathbf{u}}_{t+k|t}). \label{eq:rollout_state}
\end{align}
At each step \(k\), the predicted observation \(\hat{\mathbf{o}}_{t+k+1|t}\) is reconstructed from \(\hat{\mathbf{x}}_{t+k+1|t}\) using the observation structure in ~\eqref{eq:observation}: the robot state and goal-relative position are already known to the robot, {while the predicted LiDAR rays are generated using the historical data of the environment, i.e., obstacles encountered thus far.}
{At each rollout step, the predicted history $\widehat{\mathcal{H}}_{t+k\mid t}$ is updated by discarding its oldest
observation-action pair and appending $(\hat{\mathbf{u}}_{t+k\mid t},\hat{\mathbf{o}}_{t+k+1\mid t})$, yielding $\widehat{\mathcal{H}}_{t+k+1\mid t}$. The resulting sequence of predicted observations $\{\hat{\mathbf{o}}_{t+k\mid t}\}_{k=0}^{H}$ is passed to the barrier critic for safety evaluation.}

\textbf{Barrier critic}:~Given the predicted observation sequence $\{\hat{\mathbf{o}}_{t+k\mid t}\}_{k=0}^{H}$, a barrier critic $h_\phi:\mathbb{R}^{n_o}\rightarrow\mathbb{R}$ evaluates horizon-level safety by mapping each predicted observation to a scalar barrier value. Operating on local observations rather than the full environmental state makes the critic directly applicable to partially observable settings. At time $t$, safety is evaluated by checking whether the DTCBF condition is satisfied along the predicted observation rollout:
\begin{equation}
    \label{eq:horizon_constraint}
    h_\phi\left(\hat{\mathbf{o}}_{t+k+1}\right)
    \geq (1-\gamma)\, 
    h_\phi\left(\hat{\mathbf{o}}_{t+k}\right),
    \quad k=0,\ldots,H-1.
\end{equation}
The per-step violation of~\eqref{eq:horizon_constraint} is $v_k(\mathbf{o}_t) = \big[\eta_v + (1-\gamma)\, h_\phi(\hat{\mathbf{o}}_{t+k|t}) - h_\phi(\hat{\mathbf{o}}_{t+k+1|t})\big]_+$, where $\eta_v \geq 0$ is a margin tightening the barrier constraint. The horizon-level violation is obtained by approximating the worst-case violation $\max_kv_k(\mathbf{o_t})$ via a smooth log-sum-exp approximation:
\begin{equation}
    \label{eq:horizon_violation}
    V({\mathbf{o}_t}) = \frac{1}{\beta} \log \left( \frac{1}{H} \sum_{k=0}^{H-1} 
    \exp\left(\beta\, v_k(\mathbf{o}_t)\right) \right), \quad \beta > 0,
\end{equation}
where $\beta \rightarrow \infty$ recovers $\max_k v_k(\mathbf{o}_t)$, and the normalization by $H$ ensures $V(\mathbf{o}_t) = 0$ when all constraints are satisfied.

\textbf{SQP-based safety teacher}: The safety teacher in Step 2 (see Fig.~\ref{fig:algo-structure-left}) operates only during training to generate corrective-action labels for the actor head. Given the horizon barrier constraints~\eqref{eq:horizon_constraint} evaluated by the barrier critic $h_\phi$, the goal is to find the sequence of residual corrections $\Delta \mathbf{U}_t = (\Delta  \mathbf{u}_{t|t}, \ldots, \Delta  \mathbf{u}_{t+H-1|t})$ that minimally modifies the nominal controller while satisfying all horizon barrier constraints. Although SQP is used only during training to generate supervision for the actor head, its candidate trajectories are propagated using the {simulator dynamics} available during training, \( \mathbf{x}_{t+k+1|t}=F( \mathbf{x}_{t+k|t}, \mathbf{u}_{t+k|t})\).  This avoids training the actor head on corrective labels generated under a different rollout distribution. {At deployment, the SQP teacher is not used; the actor head predicts the residual correction directly from the observation-action history \(\mathcal{H}_t\).} The CBF constraints are then enforced on the resulting predicted observations $\hat{\mathbf{o}}_{t+k|t}$, leading to the following receding horizon optimization problem:
\begin{equation}
    \label{eq:sqp_problem}
    \min_{\Delta \mathbf{U}_t} \sum_{k=0}^{H-1} \left\| \mathbf{u}_{t+k|t} - 
    \mu_{\mathrm{nom}}(\mathbf{x}_{t+k|t}) \right\|^2 \quad
    \text{s.t.} \quad \mathbf{u}_{t+k|t} \in \mathcal{U},\ c_k(\Delta \mathbf{U}_t) \leq 0,\ 
    k = 0, \ldots, H-1,
\end{equation}
where {$c_k(\Delta \mathbf{U}_t) = \eta_s + (1-\gamma)\, h_\phi(\hat{\mathbf{o}}_{t+k|t}) - h_\phi(\hat{\mathbf{o}}_{t+k+1|t})$} encodes the DTCBF constraint \eqref{eq:horizon_constraint} at step $k$, where 
$\eta_s \geq 0$ is a safety margin that tightens the barrier constraint to account for the mismatch in the ground truth observations ($\mathbf{o}_{t+k|t}$) and predicted observations ($\hat{\mathbf{o}}_{t+k|t}$). Problem~\eqref{eq:sqp_problem} is nonconvex since $c_k$ is a nonconvex function of $\Delta \mathbf{U}_t$
Hence, we solve~\eqref{eq:sqp_problem} using sequential quadratic programming (SQP), which iteratively linearizes the nonlinear constraints and solves a sequence of QP subproblems~\cite{boggs1995sequential}. At each SQP iteration $i$, the constraints are linearized around the current correction sequence $\Delta \mathbf{U}^{(i)}_t$:
\begin{equation}
    \label{eq:sqp_linearization}
    c\left(\Delta \mathbf{U}^{(i)}_t + \delta\right) \approx c\!\left(\Delta \mathbf{U}^{(i)}_t\right) + J^{(i)}\delta, \quad  J^{(i)} = \frac{\partial c}{\partial \Delta \mathbf{U}_t}\bigg|_{\Delta \mathbf{U}^{(i)}_t}.
\end{equation}
A slack variable $\xi \in \mathbb{R}^H$ with quadratic penalty weight $\lambda_\xi > 0$ is introduced to ensure feasibility when the barrier constraints cannot be simultaneously satisfied.
\begin{equation}
\label{eq:sqp_subproblem}
\begin{aligned}
\min_{\delta,\,\xi}\;\; & \frac{1}{2}\sum_{k=0}^{H-1}
   \bigl\|\Delta\mathbf{u}^{(i)}_{t+k|t} + \delta_k\bigr\|^{2}
   + \frac{\lambda_\xi}{2}\,\|\xi\|_2^{2} \\[2pt]
\text{s.t.}\;\; &
   c^{(i)} + J^{(i)}\delta \le \xi,\quad
   \xi \ge 0,\quad
   \mu_{\text{nom}}\bigl(\hat{\mathbf{x}}^{(i)}_{t+k|t}\bigr)
   + \Delta\mathbf{u}^{(i)}_{t+k|t} + \delta_k \in \mathcal{U},
   \quad k = 0,\dots,H-1.
\end{aligned}
\end{equation}
Consistent with the receding-horizon principle, only the first element $\Delta\mathbf{u}^\star_t=[\Delta\mathbf{U}^\star_t]_0$ is retained as the supervision target for the actor head at each training step. 

\textbf{Data collection and labeling}:~
\label{sec:data-labeling}
Training data are collected on-policy by periodically executing the learned policy $\pi_\theta$. Rollout data are stored in a replay buffer as sequences of transitions, while a separate unsafe replay buffer retains rollouts containing at least one collision. During training, a batch of transitions is sampled from these buffers. Each sampled transition contains its observation--action history $\mathcal H$, whose final observation is denoted by $\mathbf o$. Let $\mathcal D_S$ and $\mathcal D_U$ denote the sets of observations labelled safe and unsafe, respectively. Although the state component $\mathbf x$ of an observation may be collision-free, the trajectory generated by the current policy may still encounter a collision at a later step. We therefore assign labels using the states that follow the observation in the same stored rollout. If $\mathbf x$ is in collision, $\mathbf o$ is added to $\mathcal D_U$. If $\mathbf{x}$ and all subsequent states over the next $T_{\mathrm{label}}$ steps are collision-free, $\mathbf{o}$ is added to $\mathcal{D}_S$; otherwise, it remains unlabelled. As $T_{\mathrm{label}}\rightarrow\infty$, these labels approach the infinite-horizon safe set under the policy used to generate the rollout, which is intractable to compute~\cite{so2024train,safety-liveness-reach-avoid,HJ-RL}. We therefore use $T_{\mathrm{label}}=32$ in all experiments. The labeling horizon $T_{\mathrm{label}}$ is separate from the transformer's predictive horizon $H$: $T_{\mathrm{label}}$ determines how many future states are considered when labeling observations, whereas $H$ determines the horizon over which barrier constraints are evaluated along predicted rollouts.

\textbf{Loss function design}:~
The transformer backbone $T_{\theta_T}$, actor head $\pi_{\theta_u}$, dynamics head $d_{\theta_x}$, and barrier critic $h_\phi$ are trained jointly by minimizing:
\begin{equation}
    \label{eq:total_loss}
    \min_{\theta_T, \theta_u, \theta_x, \phi} \lambda_{\text{act}} 
    L_{\text{act}}(\theta_T, \theta_u) + \lambda_{\text{dyn}} 
    {L}_{\text{dyn}}(\theta_T, \theta_x) +
    {L}_{\text{bar}}(\theta_T, \theta_u, \phi),
\end{equation}
where $\lambda_{\text{act}}, \lambda_{\text{dyn}}> 0$ and $\mathbf{z}_t = T_{\theta_T}(\mathcal{H}_t)$ denotes the latent representation.

\textit{Actor Loss}: The actor loss
    ${L}_{\text{act}}(\theta_T, \theta_u) = 
    \sum_{\mathcal{H} \in \mathcal{D}} \left\| 
\pi_{\theta_u}(\mathbf{z}_t) - \Delta \mathbf{u}^\star_t \right\|^2_2$  supervises the residual correction $\Delta \mathbf{u}_t = \pi_{\theta_u}(\mathbf{z}_t)$ using the first action of the safety teacher solution $\Delta \mathbf{u}^\star_t$, computed over all states in $\mathcal{D}$ for which the SQP optimization converges.

\textit{Dynamics loss}:~ The dynamics loss supervises the state-increment model $d_{\theta_x}$ using observed transitions from the simulator. Let  $\tau_t = (\mathcal{H}_t, \mathbf{x}_t, \mathbf{u}_t, \mathbf{x}_{t+1})$ denote a transition tuple and define the loss function as
    ${L}_{\text{dyn}}(\theta_T, \theta_x) = 
    \sum_{\tau_t \in \mathcal{D}} \left\| 
    d_{\theta_x}(\mathbf{z}_t,\mathbf{u}_t) - (\mathbf{x}_{t+1} - \mathbf{x}_t) \right\|^2_2$.

\textit{Barrier loss}:~The barrier loss penalizes DTCBF violations along predicted rollouts and trains $h_\phi$ to separate safe and unsafe {observations}. Using the horizon-level violation $V(\mathbf{o})$ defined in~\eqref{eq:horizon_violation}, the barrier loss combines the rollout loss and classification loss:
\begin{equation}
    \label{eq:barrier_loss}
    {L}_{\text{bar}}(\theta_T,\theta_u,\phi)
    =
    \underbrace{
    {\lambda_{\text{bar}}}
    \frac{1}{|\mathcal{D}|}
    \sum_{\mathbf{o} \in\mathcal{D}}V(\mathbf{o})
    }_{{L}_{\text{roll}}}
    +
    \underbrace{
    {\frac{\lambda_{\text{safe}}}{|\mathcal{D}_S|}}
    \sum_{\mathbf{o}\in\mathcal{D}_S}
    [\eta-h_\phi(\mathbf{o})]_+
    +
    {\frac{\lambda_{\text{unsafe}}}{|\mathcal{D}_U|}}
    \sum_{\mathbf{o}\in\mathcal{D}_U}
    [\eta+h_\phi(\mathbf{o})]_+
    }_{{L}_{\text{cls}}}.
\end{equation}
where {$\lambda_{\text{bar}} > 0$ weights the rollout loss and $\lambda_{\text{safe}}, \lambda_{\text{unsafe}} > 0$ weight the safe and unsafe classification terms}, while $\eta>0$ defines the classification margin, encouraging $h_\phi(\mathbf{o})\geq\eta$ for safe observations and $h_\phi(\mathbf{o})\leq-\eta$ for unsafe observations, preventing the barrier critic from collapsing to a trivial solution near zero. Here, $\mathcal D$ denotes the batch of histories sampled from the replay buffers. For each $\mathbf o\in\mathcal D$, the BarrierFormer generates a predicted rollout, and $V(\mathbf o)$ evaluates its horizon-level barrier violation according to~\eqref{eq:horizon_violation}. The sets $\mathcal D_S$ and $\mathcal D_U$ contain the observations labelled safe and unsafe, respectively. BarrierFormer is trained in two phases. In Phase~1, the transformer backbone $T_{\theta_T}$ and dynamics head $d_{\theta_x}$ are pretrained using ${L}_{\text{dyn}}$, while the actor head $\pi_{\theta_u}$ is initialized via behavioral cloning on offline collected trajectories~\cite{pact, conbat, transformermpc, gcbfplus}, allowing all components to warm-start before online training begins. In Phase 2, the full training objective~\eqref{eq:total_loss} is minimized with on-policy data collection; the dynamics head $d_{\theta_x}$ is updated solely via ${L}_{\text{dyn}}$, while CBF gradients flow through the barrier critic $h_\phi$ directly into the actor head $\pi_{\theta_u}$.
\section{Experiments}
\label{sec:experiments}
{We evaluate BarrierFormer to answer the following questions: (1)  Does it achieve higher safety and success rates than the current state-of-the-art without sacrificing inference latency?; (2) Does training the barrier critic with the learned dynamics head maintain performance compared to 
training with the true simulator?; and (3) Does it generalize well to out-of-distribution data, e.g., to larger and more cluttered environments?}


\textbf{Environments}: We evaluate BarrierFormer on three dynamical systems: one linear (Double Integrator) and two nonlinear (Dubins Car and Crazyflie). The sensing radius is set to $R = 0.5$ for Double Integrator and Dubins Car and $R = 1.0$ for Crazyflie, and the robot radius to $r = 0.05$ across all environments. Training is conducted in a workspace with sidelength $l=4$ and $M=8$ obstacles for Double Integrator and Dubins Car, and $l=3$ and $M=6$ obstacles for Crazyflie. The predictive horizon, observation-action history length, and labeling horizon are set to \(H=6\), \(T=12\), and \(T_{\mathrm{label}}=32\), respectively. Further details are given in Appendix~A.

\textbf{Evaluation Metrics}: We evaluate each method using three metrics. The \textbf{safety rate} is defined as the fraction of episodes in which the robot does not collide with any obstacle at any timestep. The \textbf{reaching rate} is defined as the fraction of episodes in which the robot reaches its goal location by the end of the episode. The \textbf{success rate} is defined as the fraction of episodes in which the robot is both safe and reaches its goal.   For each environment, performance is evaluated across 3 random seeds and 32 experiments, reporting mean and standard deviation.

\textbf{Baselines}: We compare BarrierFormer against four baselines based on three different model-free and one model-based paradigms:\footnote{\textit{Implementation Note.} CBF-RL and CoBL-Diffusion were originally evaluated on single integrator systems.  To enable fair comparison across all three environments, all baselines are re-implemented within our framework using the same dynamics, observation, and obstacle configurations (refer Appendix).}

\textit{RL-based}:~{CBF-RL}~\cite{cbfrl} integrates CBFs into RL training by safety filtering policy rollouts and incorporating a barrier-inspired reward term, but relies on predefined CBFs.

\textit{Diffusion-based}:~{CoBL-Diffusion}~\cite{cobl-diffusion} guides the denoising process of a diffusion model using predefined CBFs via gradient projection, enforcing safety iteratively at inference time. 

\textit{Transformer-based}:~{ConBaT}~\cite{conbat}  augments a causal transformer with a learned barrier critic trained on safe and unsafe 
demonstrations, and at inference applies a lightweight gradient-based optimization to correct the action when the future state 
is predicted to fall below a safety threshold.

\textit{Model-based MPC}~\cite{sathya2018embedded} enforces safety over a receding horizon by solving a constrained optimization problem at each control step, but requires accurate dynamics models and full state estimates, making real-time deployment computationally expensive.
\vspace{-10pt}
\begin{table}[h]
\centering
\caption{Performance comparison in terms of safety rate (Safe), goal reaching rate (Reach), success rate (Success) and per-step computational time (Time) on 2D environments 
DI: Double Integrator, DC: Dubins Car, and 3D environment CF: Crazyflie drone.  $^\dagger$Not reported for CoBL-Diffusion, whose iterative denoising makes inference slow~\cite{diffusion-speed}.}
\label{tab:results}
\resizebox{\textwidth}{!}{%
\small
\begin{tabular}{llcccc|cc}
\toprule
\multirow{2}{*}{\textbf{Env.}} & \multirow{2}{*}{\textbf{Metric}} & \multicolumn{4}{c|}{\textbf{Model-Free}} & \multicolumn{2}{c}{\textbf{Model-Based}} \\
\cmidrule(lr){3-6} \cmidrule(lr){7-8}
&                       & \textbf{CBF-RL}   & \textbf{CoBL-Diff} & \textbf{ConBaT}          & \textbf{Ours (Dyn-Head)}      & \textbf{MPC}                              & \textbf{Ours (Sim)} \\
\midrule
\multirow{3}{*}{DI} 
    & Safe              & $67.71\pm2.95$    & $76.04\pm3.90$    & $72.9\pm8.20$              & $\mathbf{100.00 \pm 0.00}$      & ${98.96\pm1.47}$               & $\mathbf{100.00 \pm 0.00}$ \\
    & Reach             & $67.71\pm2.95$    & $26.04\pm7.80$    & ${94.8\pm1.50}$     & $\mathbf{97.92 \pm 3.61}$               & ${96.86\pm2.55}$               & $\mathbf{97.92 \pm 3.61}$ \\
    & Success           & $67.71\pm2.95$    & $21.88 \pm 5.10$  & $68.8 \pm 6.80$            &  $\mathbf{97.92 \pm 3.61}$    & ${95.83 \pm 1.47}$             &  $\mathbf{97.92 \pm 3.61}$    \\
    & Time ($\mu$s)     & $9.24 \pm 0.36$   & $-^\dagger$       & $6.63 \pm 0.04$           & $\mathbf{4.99 \pm 0.02}$      & $8256 \pm 900$                        & $\textbf{4.96} \pm \textbf{0.03}$ \\
\midrule
\multirow{3}{*}{DC} 
    & Safe            & $56.25 \pm 5.10$    & $70.83\pm3.90$    & $64.6\pm13.10$             & $\mathbf{97.92 \pm 1.80}$   & $\mathbf{97.92\pm1.47}$ & $\mathbf{97.92\pm 1.80}$ \\
    & Reach           & $52.08 \pm 3.90$    & $2.08\pm2.95$     & $27.1\pm3.90$              & $\mathbf{95.83 \pm 1.80}$   & $\mathbf{95.83\pm\textbf1.47}$ & $94.79 \pm 3.61$ \\
    & Success         & $52.08 \pm 3.90$    & $2.08 \pm 2.95$   & $21.9 \pm 2.60$            & $\mathbf{93.75 \pm 0.00}$          & $\mathbf{94.79 \pm 1.47}$ & $\mathbf{94.79 \pm 3.61}$   \\
    & Time ($\mu$s)   & $9.90 \pm 0.21$     & $-^\dagger$       & $6.70 \pm 0.02$ & $\textbf{4.79} \pm \textbf{0.01}$ & $9280 \pm 1262$ & $\textbf{4.82} \pm \textbf{0.04}$ \\
\midrule
\multirow{3}{*}{CF} 
    & Safe              & $88.54\pm2.95$    & $\mathbf{98.96\pm 1.80}$ & $91.7 \pm 5.90$         & $94.79 \pm 3.61$   & ${93.75\pm0.00}$ & $\textbf{95.83}\pm \textbf{4.77}$ \\
    & Reach             & $85.42\pm 3.90$   & $0.00 \pm 0.00$           & $\mathbf{96.9\pm2.60}$ & $94.79 \pm 6.51$    & $\mathbf{100.00\pm0.00}$ & $93.75\pm6.25$ \\
    & Success           & $85.42 \pm 3.90$  & $0.00 \pm 0.00$           & $88.5 \pm 3.90 $       & $\mathbf{89.58 \pm 10.05}$  & $\mathbf{93.75 \pm 0.00}$  &  $89.58 \pm 10.05$  \\
    & Time ($\mu$s)     & $9.51 \pm 0.38$   & $-^\dagger$               & $8.63 \pm 0.03$       & $\mathbf{5.12\pm0.04}$    & $2773 \pm 103$   & $\textbf{5.08}\pm \textbf{0.03}$\\
\bottomrule
\end{tabular}}
\end{table}

We evaluate two variants of BarrierFormer that differ in how predictive rollouts are generated during training. In \textbf{Ours (Sim)}, the barrier 
critic $h_\phi$ is trained using rollouts generated by the true simulator, providing accurate state predictions for CBF constraint evaluation. In  \textbf{Ours (Dyn-Head)}, the barrier critic is trained using rollouts generated by the learned dynamics head $d_{\theta_x}$, which is jointly  trained with the transformer backbone. Table~\ref{tab:results} compares BarrierFormer against baselines across three dynamical systems, evaluated over 32 environments and 3 seeds for 256 timesteps on the training distribution. Among model-free methods, \textbf{Ours (Dyn-Head)} achieves the best balance between safety and goal-reaching with competitive inference time.  CoBL-Diffusion achieves higher safety rates on CF but at the cost of near-zero reaching rates, consistent with the local trap problem identified in~\cite{safediffuser}, where CBF-guided gradient projection during denoising can prevent the planner from making progress toward the goal. The higher performance on Crazyflie across all methods is attributed to the lower obstacle density in the 3D environment. Among model-based methods, \textbf{Ours (Sim)} matches or exceeds MPC's safety rate across all environments while reducing per-step computation time by two to three orders of magnitude. 

\begin{figure}[t]
  \centering
  \includegraphics[width=1\linewidth]{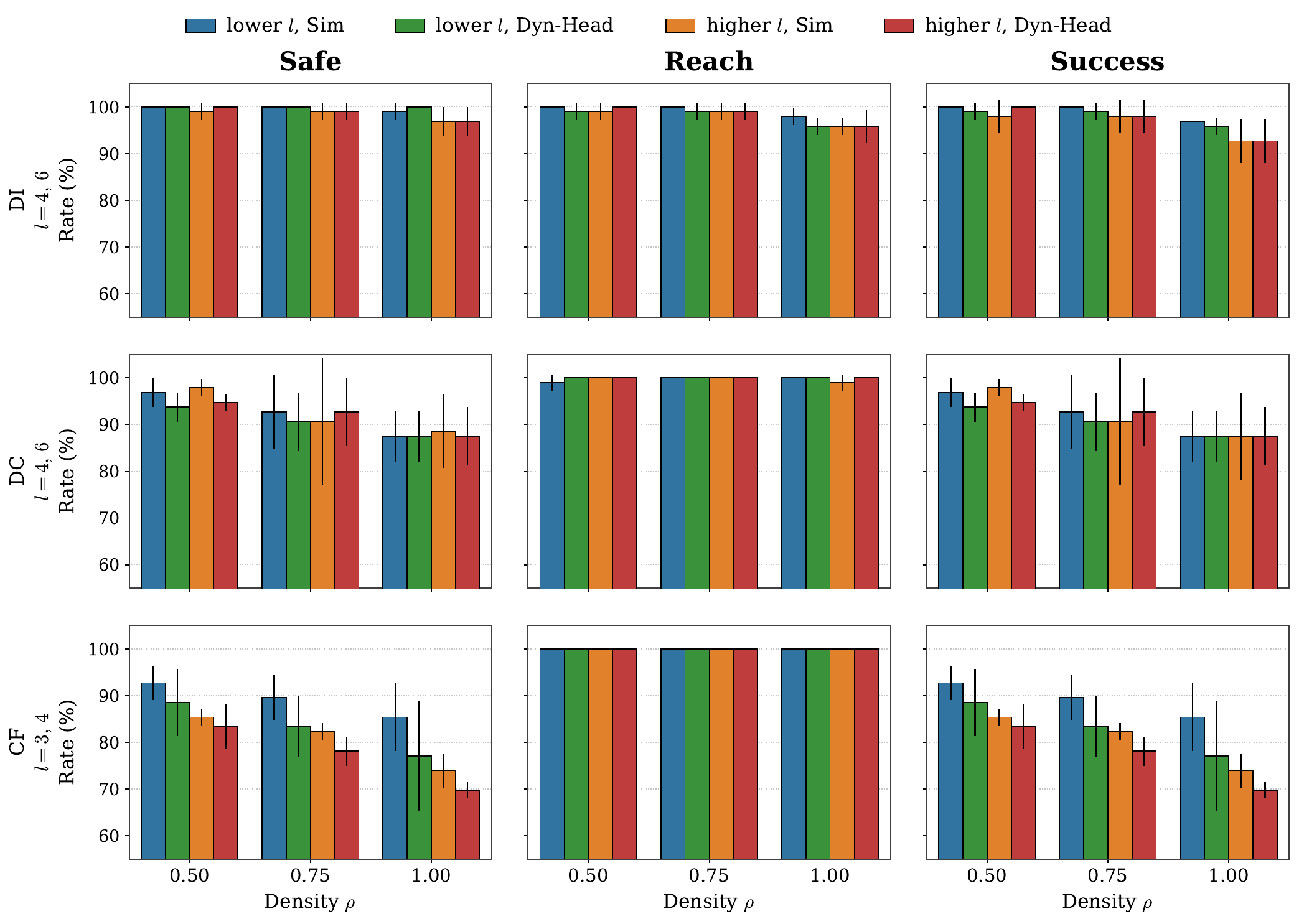}
  \caption{Zero-shot generalization across obstacle densities $\rho=M/l^{n_p}$ and workspace sizes. Rows show DI, DC, and CF; columns show safety, reaching, and success rates. Colors distinguish Sim and Dyn-Head at the smaller and larger workspace sizes. Bars report mean $\pm$ standard deviation over 32 environments and 3 seeds. The DI and DC models are trained only at $(\rho,l)=(0.50,4)$, while the CF model is trained only at $(\rho,l)\approx(0.22,3)$.}
  \label{fig:generalization}
\end{figure}

\textbf{Generalization Evaluation.} Fig.~\ref{fig:generalization} evaluates zero-shot generalization across different obstacle densities and workspace sizes. Obstacle density is defined as $\rho=M/l^{n_p}$, where $M$ is the number of obstacles and $n_p\in\{2,3\}$ is the dimension of the position space. We evaluate $\rho\in\{0.50,0.75,1.00\}$ with workspace sizes $l\in\{4,6\}$ for DI and DC and $l\in\{3,4\}$ for CF. The DI and DC models are trained at $\rho=0.50$ with $l=4$, while the CF model is trained with $M=6$ obstacles and $l=3$, corresponding to $\rho\approx0.22$. To ensure a controlled, paired comparison, each instance is initialized at the highest density, $\rho=1.00$. Lower-density variants are then obtained by removing a random subset of obstacles while preserving the poses of all remaining obstacles. Safety rates generally decrease with increasing density, as higher clutter reduces navigable space and increases the frequency of obstacle encounters. At a fixed obstacle density, performance generally decreases as the workspace size increases, potentially due to longer navigation paths and increased exposure to obstacles. Despite being trained only at $(\rho,l)=(0.50,4)$ for DI and DC and $(\rho,l)\approx(0.22,3)$ for CF, both variants retain substantial safety under the evaluated distribution shifts.

\section{Limitations}
While the proposed architecture performs better than a variety of existing prediction-based frameworks, its performance drops as problem complexity increases (from linear to nonlinear and from 2D to 3D). The resulting architecture is model-free at inference; however, the training mechanism still requires access to a black-box simulator for data collection. 
Making the framework truly model-free during training remains an open direction, requiring end-to-end online training to remove the simulator dependency and distributional bias introduced by offline pretraining. Beyond single-agent navigation, a natural extension is to multi-agent settings, where graph-based transformers could enable distributed horizon-level barrier enforcement across agents. 
Finally, similar to the existing black-box simulator-based approaches, the proposed architecture can have a large simulation-to-real-world (sim2real) gap, which might lead to undesirable behavior on real robotic systems. 
\section{Conclusion}
\label{sec:conclusion}
In this paper, we presented BarrierFormer, a barrier-supervised transformer framework that encodes rollout-level CBF constraints directly into policy parameters via training-time supervision from an SQP-based safety teacher, enabling predictive safety enforcement without online optimization at inference. {Under partial observability, BarrierFormer uses an observation--action history to generate a predictive rollout and evaluate DTCBF violations along it.} Evaluations on linear and nonlinear, 2D and 3D dynamical systems illustrate that BarrierFormer achieves better performance than model-free baselines and competitive performance with model-based MPC, while reducing inference latency by over three orders of magnitude. 

\clearpage
\acknowledgments{The Authors acknowledge the National Artificial Intelligence Research Resource (NAIRR) Pilot and AWS under grant NAIRR250034 for contributing to this research result. Any opinions, findings, conclusions, or recommendations expressed in this publication are those of the authors and do not necessarily reflect the views of the sponsors.}


\bibliography{example}  
\newpage

\appendix
\section*{\centering \Large{Appendix}}
\section{Implementation and Experiment Details} \label{sec:implementation} We implement the framework in JAX~\citep{bradbury2018jax} with the Flax neural network library, training all models end-to-end with the AdamW optimizer~\citep{loshchilov2018decoupled}. The same network architecture, optimizer, and loss formulation are used across all three environments (Double Integrator, Dubins Car, and Crazyflie); only the input and output dimensions of the observation, state, and action spaces change (refer Table~\ref{tab:env-dims}). The policy follows a learned receding-horizon design: a causal transformer encodes the robot's recent observation-action history, a residual actor head corrects a nominal controller, and a learned dynamics head provides a differentiable one-step predictor used to roll the policy forward during training. Safety is enforced by a neural barrier critic supervised by safe and unsafe labels and a discrete-time CBF condition evaluated along transformer-predicted rollouts. An SQP-based safety teacher provides imitation targets for the actor head. All weights use Xavier-uniform initialization and all environments are integrated with a fixed timestep $\Delta t = 0.03\,\mathrm{s}$. All policies are trained for $1000$ steps, with $256$ timesteps of environment interaction per step.

\subsection{Observation Representation} \label{sec:obs} For each environment, the robot observation $\mathbf{o}_t \in \mathbb{R}^{n_o}$ concatenates the robot state $\mathbf{x}_t \in \mathbb{R}^{n_x}$, the relative goal vector $\mathbf{p}_{\mathrm{goal}}-\mathbf{p}_t \in \mathbb{R}^{n_p}$, and the flattened LiDAR-ray features $\mathbf{y}_{t,j}$. The 2D environments (Double Integrator and Dubins Car) use four features per LiDAR ray: a hit indicator, a normalized distance, and the ray-direction components $(\cos\theta_j,\sin\theta_j)$. The 3D Crazyflie environment uses five features per LiDAR ray: a hit indicator, a normalized distance, and the ray-direction components $(\cos\theta_j\cos\phi_j,\cos\theta_j\sin\phi_j,\sin\theta_j)$. {In 2D, $\theta_j$ denotes the planar angle of the $j$-th ray. In 3D, $\theta_j$ denotes its elevation from the horizontal plane, while $\phi_j$ denotes its azimuth in the horizontal plane measured from the positive $x$-axis.} The resulting observation dimensions for each environment are summarized in Table~3.

\subsection{Causal Transformer Backbone}
\label{sec:transformer}
The history encoder is a GPT-2-style~\citep{radford2019language} decoder-only transformer with pre-LayerNorm (LN) residual blocks. The transformer backbone takes the observation-action history $\mathcal{H}_t$ as input and produces a latent summary ${\mathbf{z}_t \in \mathbb{R}^{d_h}}$ used by the downstream actor and dynamics heads. The history contains ${T} = 12$ observations and ${T-1}$ actions, tokenized into an interleaved sequence of length ${2T-1=23}$:
\[
\big[\,{\mathbf{o}_{t-T+1}},\,{\mathbf{u}_{t-T+1}},\,\ldots,\,{\mathbf{o}_{t-1}},\,{\mathbf{u}_{t-1}},\,{\mathbf{o}_t}\,\big].
\]
Each observation and action token is linearly projected to the model width ${d_h}$ by two separate embedding layers, and a learned absolute positional embedding is added. The sequence is processed by a single GPT-2-style~\citep{radford2019language} transformer block with 2 attention heads under a causal attention mask, consisting of pre-LayerNorm multi-head self-attention (MHSA) followed by a position-wise feed-forward network with GELU activation, both with residual connections. After the final LayerNorm, the latent at the last token position (corresponding to the current observation ${\mathbf{o}_t}$) is extracted as ${\mathbf{z}_t}$. The full configuration is summarized in Table~\ref{tab:transformer}.

\subsection{Actor Head}
\label{sec:action-head}
The actor head $\pi_{\theta_u}$ maps the transformer latent representation $\mathbf{z}_t \in \mathbb{R}^{d_h}$ to a residual correction $\Delta\mathbf{u}_t=\pi_{\theta_u}(\mathbf{z}_t)\in\mathbb{R}^{n_u}$ that is added to the action produced by the nominal control policy $\mu_{\mathrm{nom}}$: $\mathbf{u}_t=\mu_{\mathrm{nom}}(\mathbf{x}_t)+\Delta\mathbf{u}_t$. The actor head applies an input LayerNorm, {four hidden layers of width 128} with $\tanh$ activations, and a linear output layer; the environment-specific layer dimensions are summarized in Table~\ref{tab:heads}. The residual correction is left unbounded by the network, and the resulting action is clipped to the environment action bounds at execution. The nominal control policy $\mu_{\mathrm{nom}}$ is an LQR controller for the Double Integrator, a PID controller for the Dubins Car, and a two-level LQR architecture for the Crazyflie drone, where the high-level controller outputs reference velocity and yaw-rate commands that are tracked by a low-level LQR controller.

\subsection{Dynamics Head} \label{sec:dynamics-head} The dynamics head predicts a state increment $\Delta\hat{\mathbf{x}}_t=d_{\theta_x}(\mathbf{z}_t,\mathbf{u}_t)\in\mathbb{R}^{n_x}$ such that $\hat{\mathbf{x}}_{t+1}=\mathbf{x}_t+\Delta\hat{\mathbf{x}}_t$. It takes the concatenated vector $[\mathbf{z}_t^\top,\mathbf{u}_t^\top]^\top\in\mathbb{R}^{d_h+n_u}$, applies an input LayerNorm followed by three hidden layers with GELU activations and a linear output layer. The environment-specific layer dimensions are summarized in Table~\ref{tab:heads}. The dynamics-head parameters $\theta_x$ are optimized solely through the dynamics loss $L_{\mathrm{dyn}}$. When the dynamics head is used to generate predictive rollouts for the barrier loss $L_{\mathrm{bar}}$, $\theta_x$ is held fixed, while gradients with respect to its inputs $\mathbf{z}_t$ and $\mathbf{u}_t$ are retained.


\begin{table}[t]
\centering
\caption{Per-environment input/output dimensions. The network architecture is identical
across environments; only these dimensions change. ``Feat./ray'' is the number of
LiDAR features per {ray}, so
${n_o = n_x + n_p + n_{\text{rays}}\cdot(\text{Feat./ray})}$.}
\label{tab:env-dims}
\begin{tabular}{lccccc}
\toprule
Environment & State $n_x$ & Action $n_u$ & LiDAR rays ${n_{\text{rays}}}$ & Feat./ray & Observation $n_o$ \\
\midrule
Double Integrator & $4$  & $2$ & $32$ & $4$ & $134$ \\
Dubins Car        & $4$  & $2$ & $32$ & $4$ & $134$ \\
Crazyflie         & $12$ & $4$ & $32$ & $5$ & $175$ \\
\bottomrule
\end{tabular}
\end{table}

\begin{table}[t]
\centering
\caption{Causal transformer backbone hyperparameters (shared across
environments).}
\label{tab:transformer}
\small
\begin{tabular}{lc}
\toprule
\textbf{Hyperparameter} & \textbf{Value} \\
\midrule
Model width ${d_h}$   & $128$ \\
Layers / heads                      & $1$ / $2$ \\
Head dimension $d_{\text{head}}$    & $64$ \\
FFN width                           & ${4 \times d_h = 512}$ \\
FFN activation                      & GELU \\
Positional embedding                & Learned, absolute \\
History length ${T}$ & $12$ \\
Token sequence length               & ${2T-1 = 23}$ \\
\bottomrule
\end{tabular}
\end{table}
\begin{table}[t]
\centering
\caption{Training and rollout hyperparameters shared across DI, DC, and CF.}
\label{tab:hyperparameters}
\small
\begin{tabular}{lc}
\toprule
\textbf{Hyperparameter} & \textbf{Value} \\
\midrule
Actor/transformer learning rate & $3\times10^{-5}$ \\
Barrier-critic learning rate & $3\times10^{-6}$ \\
$\lambda_{\mathrm{act}}$ & $0.5$ \\
$\lambda_{\mathrm{bar}}$ & $0.5$ \\
$\lambda_{\mathrm{dyn}}$ & $1.0$ \\
$\lambda_{\mathrm{safe}}$ & $1.0$ \\
$\lambda_{\mathrm{unsafe}}$ & $1.5$ \\
{$\gamma$ (class-$\mathcal{K}$ rate)} & $0.3$ \\
{$\eta_s$ (SQP safety margin)} & $0.02$ \\
{$\eta_v$ (rollout-loss margin)} & $0.04$ \\
{$\beta$ (log-sum-exp temperature)} & $20.0$ \\
{$\eta$ (classification margin)} & $0.02$ \\
History length {$T$} & $12$ \\
Prediction/SQP horizon $H$ & $6$ \\
$\tau_{\mathrm{tgt}}$ & $0.5$ \\
$T_{\mathrm{label}}$ & $32$ \\
\bottomrule
\end{tabular}
\end{table}

\begin{table}[t]
  \centering
  \caption{Prediction head layer dimensions (${d_h} = 128$). Hidden activations are $\tanh$ for the actor head and GELU for the 
  dynamics head; output layers are linear.}
  \label{tab:heads}
  \begin{tabular}{lcc}
    \toprule
    \textbf{Environment} & \textbf{Actor head} (${d_h} \to n_u$) & 
    \textbf{Dynamics head}(${d_h+n_u} \to n_x$) \\
    \midrule
    Double Integrator & $128 \to 128 \to 128 \to 128  \to 128 \to 2$ & $130 \to 128 \to 128 \to 128 \to 4$ \\
    Dubins Car        & $128 \to 128 \to 128 \to 128  \to 128 \to 2$ & $130 \to 128 \to 128 \to 128 \to 4$ \\
    Crazyflie         & $128 \to 128 \to 128  \to 128  \to 128 \to 4$ & $132 \to 128 \to 128 \to 128 \to 12$ \\
    \bottomrule
  \end{tabular}
\end{table}

\subsection{Barrier Critic Architecture} \label{sec:cbf} The barrier critic $h_\phi:\mathbb{R}^{n_o}\to(-1,1)$ is a feed-forward network with an input LayerNorm, hidden layers of widths $(256,256,128)$ with ReLU activations, and a scalar output layer followed by a $\tanh$ activation. Since the SQP safety teacher generates supervision labels by evaluating barrier constraints along predicted rollouts, using the concurrently updated critic could produce inconsistent labels across inner training epochs. Therefore, a target critic $h_{\bar{\phi}}$ is maintained using the Polyak update $\bar{\phi}\leftarrow\tau_{\mathrm{tgt}}\phi+ (1-\tau_{\mathrm{tgt}})\bar{\phi}$, where $\tau_{\mathrm{tgt}}=0.5$, and the SQP teacher uses $h_{\bar{\phi}}$ instead of $h_\phi$ during label generation. This keeps the supervision targets fixed across the corresponding inner training epochs while the current critic $h_\phi$ is updated through the barrier loss $L_{\mathrm{bar}}$. The learning rates and loss coefficients for each environment are provided in Table~\ref{tab:hyperparameters}.

\subsection{Environment Dynamics}
\label{app:env}

\noindent\textbf{Double Integrator.} The state of the agent is given by ${\mathbf{x}} = [p_x, p_y, v_x, v_y]^\top$, where $[p_x, p_y]^\top$ is the position and $[v_x, v_y]^\top$ is the velocity. The control input is ${\mathbf{u}} = [a_x, a_y]^\top$, representing acceleration. The continuous-time dynamics are:
\begin{equation}
    {\dot{\mathbf{x}}} = \begin{bmatrix} 0 & 0 & 1 & 0 \\
    0 & 0 & 0 & 1\\
    0 & 0 & 0 & 0 \\
    0 & 0 & 0 & 0 \end{bmatrix}{\mathbf{x}} + \begin{bmatrix}
        0 & 0\\
        0 & 0\\
        1 & 0 \\
        0 & 1
    \end{bmatrix}{\mathbf{u}},
\end{equation}
discretized with timestep $\Delta t = 0.03$s. The control input is boundedcomponentwise as $\mathbf{u} \in [-1,1]^2$.

\noindent\textbf{Dubins Car.} The state of the agent is given by ${\mathbf{x}} = [p_x, p_y, \theta, v]^\top$, where $[p_x, p_y]^\top$ is the position, $\theta$ is the heading angle, and $v$ is the speed. The control input is $\mathbf{u} = [\omega, a]^\top$, containing angular velocity and longitudinal acceleration. The continuous-time dynamics are:
\begin{equation}
    {\dot{\mathbf{x}}} = \begin{bmatrix} v\cos(\theta) \\ v\sin(\theta) \\ 
    0 \\ 0 \end{bmatrix} + \begin{bmatrix}
        0 & 0\\
        0 & 0\\
        1 & 0\\
        0 & 1
    \end{bmatrix}{\mathbf{u}},
\end{equation}

discretized with timestep $\Delta t = 0.03\,\mathrm{s}$. The control input is bounded componentwise as $\mathbf{u} \in [-3,3]^2$. {Since the heading angle $\theta$ is unbounded, the agent-state part of the Dubins Car observation uses the bounded encoding $[p_x,\, p_y,\, v\cos\theta,\, v\sin\theta]^\top$ in place of $\mathbf{x}$.}

\noindent\textbf{Crazyflie Drone.} The Crazyflie quadrotor is modeled using 6-DOF rigid body dynamics~\cite{budaciu-cf} with state ${\mathbf{x}}\in\mathbb{R}^{12}$ consisting of positions, velocities, Euler angles, and angular velocities, and control input ${\mathbf{u}=[U_1,U_2,U_3,U_4]^\top}\in\mathbb{R}^4$ consisting of thrust $U_1$ and moments $U_2,U_3,U_4$. The translational dynamics are:
\begin{subequations}
\begin{align}
    \dot{p}_x &= \big(c(\phi)c(\psi)s(\theta) + s(\phi)s(\psi)\big)w - \big(s(\psi)c(\phi) - c(\psi)s(\phi)s(\theta)\big)v + uc(\psi)c(\theta) \\
    \dot{p}_y &= \big(s(\phi)s(\psi)s(\theta) + c(\phi)c(\psi)\big)v - \big(c(\psi)s(\phi) - s(\psi)c(\phi)s(\theta)\big)w + us(\psi)c(\theta) \\
    \dot{p}_z &= wc(\phi)c(\theta) - us(\theta) + vs(\phi)c(\theta)\\
    \dot u &= rv - qw + gs(\theta)\\
    \dot v & = pw - ry -gs(\phi)c(\theta)\\
    \dot w & = qu - pv -g c(\theta)c(\phi) + \frac{U_1}{m},
\end{align}
\end{subequations}
and the rotational dynamics are:
\begin{subequations}
\begin{align}
    \dot{\psi} &= r\frac{c(\phi)}{c(\theta)} + q\frac{s(\phi)}{c(\theta)} \\
    \dot{\theta} &= qc(\phi) - rs(\phi) \\
    \dot{\phi} &= p + rc(\phi)t(\theta) + qs(\phi)t(\theta) \\
    \dot{p} &= \frac{1}{I_{xx}}\big(U_4 - qr(I_{zz} - I_{yy})\big) \\
    \dot{q} &= \frac{1}{I_{yy}}\big(U_3 - pr(I_{xx} - I_{zz})\big) \\
    \dot{r} &= \frac{1}{I_{zz}}\big(U_2 - pq(I_{yy} - I_{xx})\big),
\end{align}
\end{subequations}
where $c(\cdot)$, $s(\cdot)$, and $t(\cdot)$ denote $\cos(\cdot)$, $\sin(\cdot)$, and $\tan(\cdot)${, respectively}; ${\mathbf{p}=[p_x,p_y,p_z]^\top}$ is the position, $(\phi,\theta,\psi)$ are the Euler angles, and $g={9.81\mathrm{m/s^2}}$ is gravitational acceleration. The motor {thrust-and-moment} mapping is:

\begin{equation}
    \begin{bmatrix}U_1\\ U_2 \\ U_3\\ U_4 \end{bmatrix} = 
    \begin{bmatrix}
    C_T & C_T & C_T & C_T\\
    -dC_T\sqrt{2} & -dC_T\sqrt{2} & dC_T\sqrt{2} & dC_T\sqrt{2}\\
    -dC_T\sqrt{2} & dC_T\sqrt{2} & dC_T\sqrt{2} & -dC_T\sqrt{2}\\
    -C_D & C_D & -C_D & C_D
    \end{bmatrix}
    \begin{bmatrix}\omega_1^2\\ \omega_2^2\\ \omega_3^2\\ \omega_4^2
    \end{bmatrix},
\end{equation}
where $\omega_i$ is the angular speed of the $i$-th motor. System parameters are: $I_{xx} = I_{yy} = 1.395 \times 10^{-5}$ kg$\cdot$m$^2$, 
$I_{zz} = 2.173 \times 10^{-5}$ kg$\cdot$m$^2$, $m = 0.0299$ kg, $C_T = 3.1582 \times 10^{-10}$ N/rpm$^2$, $C_D = 7.9379 \times 10^{-12}$ 
N/rpm$^2$, and $d = 0.03973$ m~\cite{budaciu-cf}. The dynamics are discretized with timestep $\Delta t = 0.03$s. The policy outputs a normalized high-level command in $[-1,1]^4$, which after scaling gives world-frame velocity targets $v_x,v_y\in[-2,2]$\,m/s, and
$v_z\in[-0.5,0.5]$\,m/s.

\subsection{Handcrafted CBF Definitions}
\label{app:cbf}

For the CBF-RL and CoBL-Diffusion baselines, we use handcrafted CBFs defined based on the agent's state and obstacle positions. All CBFs are designed to certify a minimum separation distance of $r$ between the agent and each obstacle.

\noindent\textbf{Double Integrator.} For a second-order system, we use a relative-degree-2 CBF. Let $h_0(\mathbf{x}) = \|\mathbf{p} - \mathbf{p}_{\text{obs}}\|^2 - 4r^2$ denote the position-based barrier, where $\mathbf{p}$ and $\mathbf{p}_{\text{obs}}$ are the agent and obstacle positions, respectively. The CBF is defined as:
\begin{equation}
    h_1(\mathbf{x}) = \dot{h}_0(\mathbf{x}) + \alpha_1 h_0(\mathbf{x}) = 2(\mathbf{p} - \mathbf{p}_{\text{obs}})^\top 
    (\mathbf{v} - \mathbf{v}_{\text{obs}}) + \alpha_1 h_0(\mathbf{x}),
\end{equation}
where $\alpha_1 = 10.0$ and $\mathbf{v}$, $\mathbf{v}_{\text{obs}}$ are the agent and obstacle velocities, respectively.

\noindent\textbf{Dubins Car.} Similarly, a relative-degree-2 CBF is used. Let $h_0(\mathbf{x}) = \|\mathbf{p} - \mathbf{p}_{\text{obs}}\|^2 - 4r^2$. The CBF is:
\begin{equation}
    h_1(\mathbf{x}) = 2(\mathbf{p} - \mathbf{p}_{\text{obs}})^\top(\mathbf{v} - \mathbf{v}_{\text{obs}}) + 
    \alpha_1 h_0(\mathbf{x}),
\end{equation}
where $\mathbf{v} = [v\cos\theta, v\sin\theta]^\top$ is the agent velocity in Cartesian coordinates and $\alpha_1 = 5.0$.

\noindent\textbf{Crazyflie.} Due to the high relative degree of the 6-DOF quadrotor dynamics, a relative-degree-3 CBF is required. Starting from $h_0(\mathbf{x}) = \|\mathbf{p} - \mathbf{p}_{\text{obs}}\|^2 - 4r^2$, we define:
\begin{align}
    h_1(\mathbf{x}) &= \dot{h}_0(\mathbf{x}) + \alpha_1 h_0(\mathbf{x}), \\
    h_2(\mathbf{x}) &= \dot{h}_1(\mathbf{x}) + \alpha_2 h_1(\mathbf{x}),
\end{align}
where $\alpha_1 = 30.0$ and $\alpha_2 = 50.0$. The time derivatives are computed analytically through the Crazyflie dynamics model. The  constraint $h_2(\mathbf{x}) \geq 0$ is enforced at each control step.
\section{Evaluation}
\label{app:eval}

\paragraph{Success criteria.}
All metrics are computed per episode using the simulator's geometric state with agent radius $r=0.05$. An episode is considered unsafe if the agent overlaps any obstacle at any {time step}, i.e., ${\mathrm{Unsafe}:=\max_t\mathrm{unsafe}_t}$. An episode is considered to have reached the goal if $|\mathbf{p}_t-\mathbf{p}_{\textrm{goal}}|<2r$ at any {time step}, i.e., ${\mathrm{Reached}:=\max_t\mathrm{finish}_t}$. The three reported metrics are:
\begin{equation}
{\mathrm{Safety}:=1-\mathrm{Unsafe}},\quad
{\mathrm{Reaching}:=\mathrm{Reached}},\quad
{\mathrm{Success}:=(1-\mathrm{Unsafe})\cdot\mathrm{Reached}}.
\end{equation}
Success requires the agent to reach the goal without any collision along the trajectory.

\paragraph{Aggregation.} For each environment, we run $S=3$ random seeds, {evaluating each seed on} $N=32$ independently {generated} environments with distinct start, goal, and obstacle layouts, {yielding} $S\times N=96$ episodes in total. Each seed yields one mean over its 32 episodes, and we report the mean and standard deviation of those 3 per-seed means.

\paragraph{Horizon Selection} 
Table~\ref{tab:ablation-horizon} studies the effect of the predictive horizon \(H\) on the Double Integrator environment using the history length \(T=12\). The \(H=1\) setting corresponds to pointwise barrier supervision without multi-step predictive rollout, and gives the lowest safety rate. Increasing the horizon from \(H=1\) to \(H=6\) improves safety from {\(84.38\%\) to \(100\%\) and success from \(76.04\%\) to \(97.92\%\)}, demonstrating the benefit of enforcing barrier constraints over predicted rollouts rather than only at the next step. {Beyond \(H=6\), safety remains at \(100\%\), while reaching and success rates vary within one standard deviation, so increasing \(H\) further yields no additional benefit. At the same time, longer horizons rely on longer autoregressive dynamics-head rollouts in the barrier rollout loss, whose prediction error grows with the horizon (Table~\ref{tab:dyn-head-t12}), and} introduce more linearized barrier constraints in the SQP subproblem, making the resulting correction labels more sensitive to linearization error and slack relaxation. We therefore use {\(H=6\), the smallest evaluated horizon at which safety saturates,} in the main experiments with \(T=12\).
\begin{table}[t]
\centering
\caption{Ablation study on predictive horizon \(H\) for Double Integrator with history length \(T=12\). Mean \(\pm\) std over 32 environments and 3 seeds.}
\label{tab:ablation-horizon}
\begin{tabular}{lcccc}
\toprule
Method & \(H\) & Safe & Reach & Success \\
\midrule
\multirow{5}{*}{Ours (Dyn-Head)}
& 1  & 84.38 $\pm$ 3.12 & 86.46 $\pm$ 7.22 & 76.04 $\pm$ 7.86 \\
& 3  & 91.67 $\pm$ 1.80 & 91.67 $\pm$ 3.61 & 78.12 $\pm$ 11.27 \\
& 6  & 100.00 $\pm$ 0.00 & 97.92 $\pm$ 3.61 & 97.92 $\pm$ 3.61 \\
& 9  & 100.00 $\pm$ 0.00 & 94.79 $\pm$ 3.61 & 94.79 $\pm$ 3.61 \\
& 12 & 100.00 $\pm$ 0.00 & 96.88 $\pm$ 3.12 & 96.88 $\pm$ 3.12 \\
\bottomrule
\end{tabular}
\end{table}
\begin{table}[t]
\centering
\caption{Dynamics-head rollout RMSE on the Double Integrator with history length $T=12$. The $H=1$ column reports the one-step RMSE; the remaining columns report the autoregressive rollout RMSE over horizons $H$.}
\label{tab:dyn-head-t12}
\small
\begin{adjustbox}{max width=\columnwidth}
\begin{tabular}{lccccc}
\toprule
\textbf{Input signal}
& $H=1$ & $H=8$ & $H=16$ & $H=24$ & $H=32$ \\
\midrule
GCBF+                                   & $0.000282$    & $0.00162$     & $0.00210$     & $0.00258$     & $0.00310$ \\
QP                                      & $0.000913$    & $0.00215$     & $0.00255$     & $0.00320$     & $0.00393$ \\
LQR                                     & $0.000383$    & $0.00143$     & $0.00186$     & $0.00253$     & $0.00301$ \\
Actor head (BarrierFormer)              & $0.000350$    & $0.00160$     & $0.00214$     & $0.00247$     & $0.00301$ \\
Actor head + noise                      & $0.000343$    & $0.00159$     & $0.00219$     & $0.00271$     & $0.00300$ \\
\midrule
Constant                                & $0.000567$    & $0.00409$     & $0.00395$     & $0.00483$     & $0.00539$ \\
Uniform random                          & $0.001910$    & $0.00431$     & $0.00514$     & $0.00507$     & $0.00489$ \\
Matched random                          & $0.001441$    & $0.00449$     & $0.00543$     & $0.00558$     & $0.00680$ \\
Polynomial                              & $0.000652$    & $0.00460$     & $0.00499$     & $0.00484$     & $0.00538$ \\
Sinusoidal                              & $0.001209$    & $0.00365$     & $0.00530$     & $0.00669$     & $0.00831$ \\
Zero                                    & $0.000204$    & $0.00447$     & $0.00667$     & $0.00881$     & $0.01113$ \\
\bottomrule
\end{tabular}
\end{adjustbox}
\end{table}


\begin{table}[t]
\centering
\caption{Dynamics-head rollout RMSE on the Double Integrator with history length $T=32$. The $H=1$ column reports the one-step RMSE; the remaining columns report the autoregressive rollout RMSE over horizons $H$.}
\label{tab:dyn-head-t32}
\small
\begin{adjustbox}{max width=\columnwidth}
\begin{tabular}{lccccc}
\toprule
\textbf{Input signal}
& $H=1$ & $H=8$ & $H=16$ & $H=24$ & $H=32$ \\
\midrule
GCBF+                                   & $0.000170$    & $0.00084$    & $0.00171$    & $0.00246$    & $0.00326$ \\
QP                                      & $0.000290$    & $0.00169$    & $0.00229$    & $0.00310$    & $0.00386$ \\
LQR                                     & $0.000130$    & $0.00110$    & $0.00176$    & $0.00252$    & $0.00325$ \\
Actor head (BarrierFormer)              & $0.000300$    & $0.00089$    & $0.00163$    & $0.00243$    & $0.00332$ \\
Actor head + noise                      & $0.000300$    & $0.00098$    & $0.00164$    & $0.00242$    & $0.00335$ \\
\midrule
Constant                                & $0.000220$    & $0.00176$    & $0.00226$    & $0.00285$    & $0.00342$ \\
Uniform random                          & $0.000620$    & $0.00194$    & $0.00217$    & $0.00251$    & $0.00292$ \\
Matched random                          & $0.000540$    & $0.00167$    & $0.00231$    & $0.00259$    & $0.00322$ \\
Polynomial                              & $0.000370$    & $0.00203$    & $0.00232$    & $0.00261$    & $0.00297$ \\
Sinusoidal                              & $0.000360$    & $0.00161$    & $0.00248$    & $0.00321$    & $0.00410$ \\
Zero                                    & $0.000150$    & $0.00484$    & $0.00776$    & $0.00961$    & $0.01154$ \\
\bottomrule
\end{tabular}
\end{adjustbox}
\end{table}

\paragraph{Pretraining.}
In Phase~1, we jointly pretrain the transformer backbone, actor head, and dynamics head using trajectories collected from three different control policies: GCBF+~\cite{gcbfplus}, a QP safety controller, and the nominal LQR controller. Together, these policies provide a diverse training distribution comprising learned safe actions, optimization-based safety corrections, and nominal control inputs. This diversity mitigates overfitting to the narrow action distribution induced by a single controller and helps preserve rollout accuracy as predictive-horizon CBF supervision guides actor updates during Phase~2. For each transition, the executed action and observed state increment provide supervision for the dynamics head. The GCBF+ and QP trajectories also provide residual-action labels for the actor head, while the nominal LQR trajectories provide additional dynamics supervision around the nominal controller. The models with history lengths \(T=12\) and \(T=32\) use the same training data and optimization procedure.

We evaluate each pretrained dynamics head on \(50\) independently generated episodes of \(256\) steps for each test input. The closed-loop inputs comprise the nominal LQR controller, the BarrierFormer actor added to the nominal action, its noise-perturbed variant with additive Gaussian noise of standard deviation \(0.1\), GCBF+, and the QP controller. The open-loop inputs are generated independently for each action dimension, with their parameters resampled for every episode. Constant inputs hold a value sampled uniformly from \([-1,1]\) throughout the episode. Uniform-random inputs are sampled independently from \(\mathcal{U}(-1,1)\) at every step, while matched-random inputs are sampled from a zero-mean Gaussian distribution with standard deviation \(0.3\) and clipped to the action bounds. Sinusoidal inputs use amplitudes sampled from \([0.3,1]\), frequencies from \(0.5\) to \(6\) cycles per episode, and phases from \([0,2\pi]\). Polynomial inputs are random cubic signals normalized to \([-1,1]\), and zero-action inputs remain zero throughout the episode. These test families span closed-loop, smooth low-frequency, and broadband random inputs, providing a range of temporal structures not encountered during pretraining. For each input signal, the dynamics head is applied autoregressively, and state RMSE is reported for prediction horizons \(H\in\{1,8,16,24,32\}\), where \(H=1\) corresponds to one-step prediction. Tables~\ref{tab:dyn-head-t12} and~\ref{tab:dyn-head-t32} present the results for \(T=12\) and \(T=32\), respectively. Increasing the history length to \(T=32\) consistently reduces prediction error on controller-based inputs through \(H=24\). At \(H=32\), the two history lengths perform comparably on these inputs, with mean RMSEs of \(0.00321\) and \(0.00341\) for \(T=12\) and \(T=32\), respectively. The longer history provides a clearer benefit on unseen nonzero open-loop inputs, reducing their mean \(H=32\) RMSE from \(0.00615\) to \(0.00333\), a reduction of approximately \(46\%\). The zero-action input is the exception, for which both models accumulate similar long-horizon error.

\paragraph{Frozen-Backbone Ablation.} To evaluate the contribution of online training of the transformer backbone and dynamics head, we evaluate a variant in which the transformer backbone and dynamics head are frozen after Phase~1 pretraining, and only the actor head and barrier critic are trained during Phase~2. The results reported in Table~\ref{tab:ablation-frozen} show degraded performance compared to the full online training setup described in Section~\ref{sec:methodology}, which we attribute to two factors. First, the frozen dynamics head accumulates prediction errors relative to the true simulator as the actor head’s action distribution shifts away from the pretraining distribution; since the transformer backbone is also frozen, the latent representation $\mathbf{z}_t$ cannot adapt to the shifting action distribution, resulting in inaccurate rollout predictions for barrier constraint evaluation. This follows the same motivation as the DAgger framework~\cite{dagger}: collecting data under the current policy reduces distribution shift, while here the resulting on-policy transitions allow the dynamics head to adapt to the policy's evolving state-action distribution. Second, freezing the backbone fixes the latent representation $\mathbf{z}_t$ to features learned in Phase~1 for dynamics prediction and behavior cloning, which are not optimized for safety. In the full setup, the actor and barrier losses in~\eqref{eq:total_loss} also update $\theta_T$, allowing the attention mechanism to adapt $\mathbf{z}_t$ to safety-relevant parts of the observation-action history; with a frozen backbone, barrier supervision can only reshape the mapping from a fixed feature space to actions. Together, these results motivate the full Phase~2 training, in which all components are updated jointly as described in Section~\ref{sec:methodology}.
\begin{table}[t]
\centering
\caption{Ablation study on freezing the transformer backbone and dynamics head after {Phase~1}. The frozen variant trains only the actor head and barrier
critic in {Phase~2}, while \emph{Ours} denotes the full training setup. {Mean $\pm$ standard deviation across 3 seeds, with each seed evaluated on 32 environments}.}
\label{tab:ablation-frozen}
\resizebox{\textwidth}{!}{%
\begin{tabular}{lcccccc}
\toprule
\multirow{2}{*}{\textbf{Env.}}
& \multicolumn{2}{c}{\textbf{Safe}}
& \multicolumn{2}{c}{\textbf{Reach}}
& \multicolumn{2}{c}{\textbf{Success}} \\
\cmidrule(lr){2-3}
\cmidrule(lr){4-5}
\cmidrule(lr){6-7}
& \textbf{Frozen} & \textbf{Ours}
& \textbf{Frozen} & \textbf{Ours}
& \textbf{Frozen} & \textbf{Ours} \\
\midrule
DI & $90.62 \pm 5.41$ & $\mathbf{100.00\pm0.00}$ & ${91.67 \pm 4.77}$ & $\mathbf{97.92 \pm 3.21}$ & $84.38 \pm 8.27$ & $\mathbf{97.92 \pm 3.21}$ \\
DC & $82.29 \pm 1.80$ & $\mathbf{97.92\pm1.80}$ & $\mathbf{96.88 \pm 3.12}$ & $95.83\pm1.80$ & $80.21 \pm 1.80$ & $\mathbf{93.75\pm0.00}$ \\
CF & $93.75\pm3.12$ & $\mathbf{94.79\pm3.61}$ & $\mathbf{95.79 \pm 6.51}$ & ${94.79\pm6.51}$ & $88.54 \pm 9.02$ & $\mathbf{89.58\pm10.05}$ \\
\bottomrule
\end{tabular}%
}
\end{table}
\section{Baseline Implementation} \label{sec:baselines} All four baselines are re-implemented in JAX/Flax within our simulation 
framework using the same dynamics, observation model, obstacle configurations, and evaluation protocol. Specifically, all methods are 
evaluated over episodes of 256 timesteps with $\Delta t = 0.03$s, on a $l=4$ workspace with 8 obstacles for DI and DC, and a $l=3$ workspace 
with 6 obstacles for CF, across 3 seeds $\times$ 32 environments.
\subsection{ConBaT} \label{app:conbat} We re-implement ConBaT~\cite{conbat} in JAX/Flax following the original formulation. ConBaT learns a safe policy from demonstrations in two stages on top of a PACT causal transformer backbone~\cite{pact}, and enforces safety at deployment through a conditional gradient-based action correction governed by a learned discrete-time control barrier function.

\paragraph{Data.} Safe trajectories are collected by rolling out an LQR reference controller, and unsafe trajectories are collected by adding Gaussian exploration noise ($\sigma=0.5$) until a collision occurs. We collect $7{,}500$ safe and $2{,}500$ unsafe trajectories of length $256$, sliced into context windows of length ${T_{\mathrm{ConBaT}}=16}$.

\paragraph{Architecture.} The PACT backbone processes an interleaved state-action token sequence with a GPT-2-style causal transformer (2 layers, embedding size 64, 8 heads, context ${T_{\mathrm{ConBaT}}=16}$), with a policy head and a world-model head. Two barrier critics are added: a current-state critic ${C(\mathbf{s}^+_t) \in \mathbb{R}}$ and a future-state critic ${C_f(\mathbf{s}^+_t,\mathbf{a}^+_t) \in \mathbb{R}}$, each a two-layer MLP with 128 hidden units and $\tanh$ activation. By convention, ${C(\mathbf{s}^+_t)>0}$ denotes safety.

\paragraph{Phase I: PACT pre-training.} The PACT backbone is pre-trained for 50 epochs (batch 32, Adam, {learning rate} $10^{-4}$) to minimize a masked action-prediction loss on safe timesteps and a world-model next-state-embedding loss weighted by $\lambda_{\mathrm{wm}}=0.1$.

\paragraph{Phase II: Barrier critic training.} With PACT frozen, the two barrier critics are trained for 10 epochs with three losses: a classification loss pushing ${C(\mathbf{s}^+_t)>\gamma_{\mathrm{ConBaT}}}$ on safe states and ${C(\mathbf{s}^+t)<-\gamma_{\mathrm{ConBaT}}}$ on unsafe states, a DTCBF forward-invariance loss enforcing ${C(\mathbf{s}^+_{t+1})\geq(1-\alpha_{\mathrm{ConBaT}})C(\mathbf{s}^+t)}$, and a consistency loss training $C_f$ to predict the next-state safety score. We use ${\gamma_{\mathrm{ConBaT}}=1.0}$, ${\alpha_{\mathrm{ConBaT}}=0.1}$, and loss weights $(\lambda_{\mathrm{c}},\lambda_{\mathrm{s}},\lambda_{\mathrm{f}})=(1,5,1)$.

\paragraph{Deployment.} At each step, the policy proposes an action $\hat{\mathbf{a}}_t$, whose action embedding $\hat{\mathbf{a}}^+_t$ is evaluated by the future-state critic. If ${C_f(\mathbf{s}^+_t,\hat{\mathbf{a}}^+_t)<0}$, the predicted next state is classified as unsafe and a gradient-descent correction is applied ($n_{\mathrm{corr}}=3$ steps, step size $0.05$, and regularization $\lambda_{\mathrm{reg}}=0.1$); otherwise, the proposed action is executed unchanged.

\subsection{CBF-RL} \label{app:cbfrl} We re-implement CBF-RL~\cite{cbfrl} in JAX/Flax following the original formulation. CBF-RL trains a PPO policy~\cite{ppo} under two complementary CBF-based safety mechanisms during training: a closed-form CBF-QP action filter and a barrier-inspired reward penalty. \paragraph{Network architecture.} The actor-critic uses a two-hidden-layer MLP with width $256$ and ReLU activations. The input concatenates the robot state, relative goal vector, and flattened obstacle encoding. Per-environment dimensions are summarized in Table~\ref{tab:cbfrl_arch}.

\begin{table}[h]
\centering
\caption{CBF-RL actor-critic input and output dimensions per environment.}
\label{tab:cbfrl_arch}
\begin{tabular}{lccc}
\toprule
\textbf{Environment} & \textbf{Input dim} & \textbf{Action dim} & 
\textbf{Obstacles} \\
\midrule
Double Integrator & 48 & 2 & 8 rectangular \\
Dubins Car        & 48 & 2 & 8 rectangular \\
Crazyflie         & 36 & 4 & 6 spherical \\
\bottomrule
\end{tabular}
\end{table}

\paragraph{PPO training.} Training uses PPO with GAE (${\gamma_{\mathrm{PPO}}=0.99}$, ${\lambda_{\mathrm{GAE}}=0.95}$, clip ${\epsilon_{\mathrm{clip}}=0.2}$), Adam optimizer ({learning rate} $3\times10^{-4}$), global-norm gradient clipping ($0.5$), 10 epochs and 8 mini-batches per update, and 256 parallel environments.

\paragraph{CBF safety filter.} The handcrafted CBFs defined in Appendix~\ref{app:cbf} are used for each environment. Writing the CBF condition as ${\psi=\mathbf{a}^{\top}\mathbf{u}\cdot{\mathrm{des}}+b}$, the closed-form filtered action is:
\begin{equation}
{\mathbf{u} =\mathbf{u}_{\mathrm{des}} -\frac{\min(\psi,0)} {\lVert\mathbf{a}\rVert_2^2}\mathbf{a}},
\end{equation}
so the action is modified only when ${\psi<0}$. A barrier penalty $10\cdot\min(h_{\min},0)$ is added to the reward at each step. At inference, the closed-form CBF filter remains active as a runtime safety layer.

\subsection{CoBL-Diffusion} \label{app} We re-implement CoBL-Diffusion~\cite{cobl-diffusion} in JAX/Flax following the original formulation. CoBL-Diffusion is a receding-horizon diffusion planner that generates a horizon-length control sequence guided at sampling time by gradient-based CBF and CLF terms.

\paragraph{Data.} We collect $50{,}000$ expert trajectories of length $256$ per environment by rolling out the reference LQR controller. Raw states are converted to ${(\mathbf{u}_t,\mathbf{p}_t)}$ pairs and normalized by the arena side length ${l=4\text{ for DI/DC and }l=3\text{ for CF}}$.

\paragraph{Model and training.} The denoiser is a 1-D temporal U-Net (base width $32$, channel multipliers $[1,2,4,8]$) that predicts the clean control sequence ${\mathbf{U}^{(0)}}$ from a noised sequence, conditioned on the diffusion time step and a position trajectory initialized as the straight start-goal interpolation. A cosine $\beta$ schedule with $1000$ diffusion steps is used. Training minimizes a three-term objective:
\begin{equation}
{ L = L_{\mathrm{simple}} + L_{\mathrm{traj}} + 10\,L_{\mathrm{term}}}.
\end{equation}
The objective consists of an $\ell_1$ denoising loss, a trajectory-reconstruction loss obtained by integrating the denoised controls through the true dynamics, and a terminal goal-reaching loss. Training uses Adam ({learning rate} $2\times10^{-5}$, global-norm clip $5.0$), batch size $512$, $1500$ epochs, and an exponential moving average (EMA) of the model weights with decay $0.995$.

\paragraph{Deployment.} At inference, the planner operates in a receding-horizon fashion: it generates an $H_{\mathrm{CoBL}}=80$-step control sequence, applies only the first control, and then replans from the resulting state. The horizon of ${H_{\mathrm{CoBL}}=80}$ steps is chosen to balance planning accuracy against the computational cost of DDIM sampling with gradient guidance at each denoising step.

\subsection{Model Predictive Control (MPC)} \label{app:mpc} We implement MPC~\cite{sathya2018embedded} as a classical model-based non-learning baseline using CasADi with the IPOPT interior-point solver. 

\paragraph{Optimal control problem.} At each control step, the MPC solves an ${H_{\mathrm{MPC}}=20}$-step trajectory optimization problem (${0.6,\mathrm{s}}$ horizon) over states ${\mathbf{X} \in \mathbb{R}^{n_x \times (H_{\mathrm{MPC}}+1)}}$ and controls ${\mathbf{U} \in \mathbb{R}^{n_u \times H_{\mathrm{MPC}}}}$. The initial state is pinned as an equality constraint, dynamics are enforced as Euler-step equality constraints, and state/control limits are imposed as box constraints. The cost combines a running and terminal quadratic position error toward the goal and a control-effort regularization term. The solution is warm-started at each step by shifting the previous solution, reducing IPOPT iterations from $\sim$50 to $\sim$5. Only the first control $u_0$ is applied at each step. 

\paragraph{Obstacle handling.} For Double Integrator and Dubins Car, rectangular obstacles are approximated by their circumscribed circles, imposing ${|\mathbf{p}_k-\mathbf{c}_i|^2 \geq (r+r_i+\epsilon)^2}$ at every node. This is conservative but smooth and always feasible. For Crazyflie, spherical obstacles admit an exact distance constraint. The safe/unsafe metric is always computed using the exact obstacle geometry, independent of the planner's conservative approximation.

\textbf{Robustness to model perturbations.} Table~11 reports the performance of BarrierFormer under dynamics perturbations not encountered during training. The policy is trained using the unperturbed Double Integrator dynamics and evaluated without retraining under two test-time perturbations. First, we introduce a parametric velocity mismatch,
\[
\dot{\mathbf{p}} = \alpha_{\mathrm{vel}} \mathbf{v},
\]
where $\alpha_{\mathrm{vel}} = 1$ corresponds to the unperturbed dynamics, and $\alpha_{\mathrm{vel}} \in [0.8, 1.2]$ changes how fast the robot moves for the same velocity state. Second, we add state-wise scaled process noise at every control step ($\Delta t = 0.03$\,s),
\[
\mathbf{x}_{t+1} = F(\mathbf{x}_t, \mathbf{u}_t) + \mathbf{w}_t, \qquad
\mathbf{w}_t \sim \mathcal{N}(\mathbf{0}, \boldsymbol{\Sigma}_w), \qquad
\boldsymbol{\Sigma}_w = \mathrm{diag}(\sigma_p^2, \sigma_p^2, \sigma_v^2, \sigma_v^2),
\]
where $F$ denotes the discretized dynamics, and $\sigma_p$ and $\sigma_v$ denote the position and velocity disturbance standard deviations, respectively. In these experiments, $\sigma_p = 4\sigma$ and $\sigma_v = 0.5\sigma$, scaled by the workspace size and the velocity limit, respectively. As shown in Table~11, under velocity mismatch, the safety rate remains at 100\% across all tested values of $\alpha_{\mathrm{vel}}$. The reaching rate decreases from 97.92\% under the unperturbed dynamics to 80.21\% at $\alpha_{\mathrm{vel}} = 0.8$, consistent with slower motion under a fixed episode time limit. Additive process noise is more challenging: the safety rate remains at or above 97.92\% for $\sigma \le 0.0025$, but decreases to 90.62\% at $\sigma = 0.005$ and 54.17\% at $\sigma = 0.01$. These results are empirical robustness tests, not formal robust safety guarantees.
\begin{table}[t]
\centering
\caption{Robustness on the Double Integrator environment under dynamics perturbations. Policies are trained using the unperturbed dynamics and evaluated without retraining. Rates are reported as {mean $\pm$ standard deviation} over 3 seeds with 32 environments per seed. For process noise, {$\sigma_p=4\sigma$ and $\sigma_v=0.5\sigma$}.}
\label{tab:robustness}
\begin{tabular}{llccc}
\toprule
\multirow{2}{*}{Perturbation} & \multirow{2}{*}{Value}
& \multicolumn{3}{c}{$T=12,\ H=6$} \\
\cmidrule(lr){3-5}
& & Safe & Reach & Success \\
\midrule
\multirow{5}{*}{$\dot{\mathbf{p}}=\alpha_{\mathrm{vel}}\mathbf{v}$}
& 0.8 & $100.00 \pm 0.00$ & $80.21 \pm 6.51$ & $80.21 \pm 6.51$ \\
& 0.9 & $100.00 \pm 0.00$  & $92.71 \pm 4.77$ &  $92.71 \pm 4.77$ \\
& 1.0 & $100.00 \pm 0.00$  & $97.92 \pm 3.61$ & $97.92 \pm 3.61$ \\
& 1.1 & $100.00 \pm 0.00$ & $100.00 \pm 0.00$ & $100.00 \pm 0.00$ \\
& 1.2 & $100.00 \pm 0.00$  & $100.00 \pm 0.00$ & $100.00 \pm 0.00$ \\
\midrule
\multirow{6}{*}{$\dot{\mathbf{x}}=f(\mathbf{x},\mathbf{u})+\mathbf{w}$}
& $\sigma=0$      & $100.00 \pm 0.00$ & $97.92 \pm 3.61$ & $97.92 \pm 3.61$ \\
& $\sigma=0.001$  & $100.00 \pm 0.00$ & $96.88 \pm 5.41$ & $96.88 \pm 5.41$ \\
& $\sigma=0.0025$ & $97.92 \pm 3.61$ & $96.88 \pm 5.41$ & $95.83 \pm 7.22$ \\
& $\sigma=0.005$  & $90.62 \pm 8.27$ & $93.75 \pm 10.83$ & $86.46 \pm 15.42$ \\
& $\sigma=0.0075$ & $75.00 \pm 6.25$ & $89.58 \pm 10.97$ & $69.79 \pm 14.77$ \\
& $\sigma=0.01$   & $54.17 \pm 9.02$ & $86.46 \pm 13.01$ & $51.04 \pm 14.43$ \\
\bottomrule
\end{tabular}
\end{table}

\section{Inference Latency Measurement} \label{app:latency} All inference latency measurements are performed on a single workstation equipped with an NVIDIA RTX 4000 Ada Generation GPU ($20\,\mathrm{GB}$), an Intel Xeon w5-3423 CPU (12 cores, up to $4.2\,\mathrm{GHz}$), and $62\,\mathrm{GiB}$ of RAM, running Ubuntu 22.04. Latency is measured as the per-step wall-clock time averaged over a full rollout of 256 steps across 32 parallel environments, with a single untimed warm-up run to exclude just-in-time (JIT) compilation cost. \texttt{jax.block\_until\_ready()} is called after every timed step to account for asynchronous dispatch. The per-step latency reported in Table~\ref{tab:results} is the total batched wall time divided by the number of steps.

\end{document}